\documentclass{article} %
\usepackage[final]{colm2026_conference}

\usepackage{microtype}
\usepackage{hyperref}
\usepackage{url}
\usepackage{booktabs}

\usepackage{lineno}

\definecolor{darkblue}{rgb}{0, 0, 0.5}
\hypersetup{colorlinks=true, citecolor=darkblue, linkcolor=darkblue, urlcolor=darkblue}

\usepackage{booktabs}
\usepackage{multirow}
\usepackage{pifont}%
\newcommand{\cmark}{\ding{51}}%
\newcommand{\xmark}{\ding{55}}%
\usepackage[svgnames]{xcolor}
\usepackage{enumitem}
\usepackage{amsmath,amssymb,stmaryrd}
\usepackage{graphicx}
\usepackage{caption}
\usepackage{subcaption}
\usepackage{xcolor} 
\usepackage{wrapfig}
\usepackage[T1]{fontenc}
\usepackage[utf8]{inputenc}

\usepackage{verbatim}

\usepackage{listings}

\definecolor{codegreen}{rgb}{0,0.6,0}
\definecolor{codegray}{rgb}{0.5,0.5,0.5}
\definecolor{codepurple}{rgb}{0.58,0,0.82}
\definecolor{backcolour}{rgb}{0.95,0.95,0.92}

\lstdefinestyle{mystyle}{
    backgroundcolor=\color{backcolour},   
    commentstyle=\color{codegreen},
    keywordstyle=\color{magenta},
    numberstyle=\tiny\color{codegray},
    stringstyle=\color{codepurple},
    basicstyle=\ttfamily\scriptsize,
    breakatwhitespace=false,         
    breaklines=true,                 
    captionpos=b,                    
    keepspaces=true,                 
    numbers=left,                    
    numbersep=5pt,                  
    showspaces=false,                
    showstringspaces=false,
    showtabs=false,                  
    tabsize=2
}
\title{Transformer See, Transformer Do: Copying as an Intermediate Step in Learning Analogical Reasoning}

\author{Philipp Hellwig$^1$, Willem Zuidema$^2$, Claire Stevenson$^1$\thanks{ shared senior authorship} \& Martha Lewis$^2$\footnotemark[1]\\
$^1$Psychological Methods, $^2$Institute for Logic, Language and Computation\\
University of Amsterdam\\
\texttt{\{p.t.hellwig, zuidema, c.e.stevenson, m.a.f.lewis\}@uva.nl}
}

\begin{document}

\ifcolmsubmission
\linenumbers
\fi

\maketitle

\newcommand{\ph}[1]{\textcolor{red}{#1}}
\newcommand{\ml}[1]{\textcolor{teal}{#1}}
\newcommand{\jz}[1]{\textcolor{blue}{#1}}
\newcommand{\cs}[1]{\textcolor{pink}{#1}}
\newcommand{\red}[1]{\textcolor{red}{#1}}
\newcommand{\blue}[1]{\textcolor{blue}{#1}}
\newcommand{\green}[1]{\textcolor{green}{#1}}

\begin{abstract}
Analogical reasoning is a hallmark of human intelligence, enabling us to solve new problems by transferring knowledge from one situation to another. Yet, developing artificial intelligence systems capable of robust human-like analogical reasoning has proven difficult. In this work, we train transformers using Meta-Learning for Compositionality (MLC) on an analogical reasoning task (letter-string analogies) and assess their generalization capabilities. We find that letter-string analogies become learnable when guiding the models to attend to the most informative problem elements, induced by including copy tasks in the training data. Furthermore, generalization to new alphabets improves when models are trained with more heterogeneous datasets. For the best training run, our 3-layer encoder-decoder model performs on par with frontier models on our letter-string analogy datasets. The MLC approach also enables some generalization to compositions of trained transformations, but not to completely novel transformations. To understand how the model solves the analogies, we identify an algorithm for one analogy task that approximates the model's computations. We verify this using interpretability analyses and show that head-level attention weights causally affect model output, allowing steering between different analogy tasks. We discuss the implications of our findings for generalization capabilities of larger models and the parallels to human analogical reasoning.
\end{abstract}

\section{Introduction}

Analogical reasoning is a process where a reasoner recognizes the \emph{relation} between objects, events or properties, and can apply that relation to new instances. For instance, successful  analogical reasoning is needed to predict ``\emph{roots}'' in the simple riddle ``\emph{body is to feet, like tree is to...?}''. 
Analogical reasoning is seen as a core component of human cognition, enabling us to transfer knowledge to new tasks and facilitating abstraction \citep{gentner2017analogy}. 

Transformer-based Large Language Models (LLMs) can solve analogical reasoning tasks to some degree \citep{webb2023emergent}; however, the abilities of LLM on such tasks are highly context dependent \citep{hodel2024response, lewis2024evaluating, stevenson2025can}, echoing results that show reduced performance on a wide range of tasks that lie outside of the training distribution \citep{mccoy2024embers, wu2024reasoning, chollet2025arc}. %
Humans, including children, can generalize to new contexts for analogical reasoning tasks \citep{brown1988preschool, doumas2022theory, stevenson2025can, lewis2024evaluating}.  
A key factor thought to be behind this human ability is that humans can use knowledge acquired from earlier tasks to learn new tasks: we are capable of \textit{meta-learning}, or learning to learn \citep{wang2021metalearning}. 
This facilitates fast and efficient learning when little information is available.

In this paper, we build on a simple but general form of analogical reasoning, as proposed by \citet{hofstadter1994copycat}, that uses letter-string analogies (e.g., ``\emph{If abc changes to abe, what should rst change to...?}''). We furthermore build on a formalization of meta-learning proposed by \citet{lake2023humanlike}, that can be seen as the route to \emph{in-context learning}. We combine both traditions to investigate whether meta-learning can improve the ability of transformer-based models to discover general solutions, that generalize in systematic ways over the alphabets and analogical patterns shown during training.

We develop a suite of datasets that tests analogical reasoning via one- and few-shot letter-string analogy problems (Section \ref{sec:data_gen}). We train small encoder-decoder transformers on this dataset (see Figure \ref{fig:arch}, details in Appendix \ref{apx:detailed_training}), and systematically evaluate how well these models learn the training task and how well they generalize to various targets, including new shuffled alphabets, new combinations of seen analogical transformations, or entirely new transformations.\footnote{Our suite of datasets can be found at \url{https://huggingface.co/datasets/philipp-hellwig/permuted-letter-string-analogies} and the code used for this paper is located at \url{https://github.com/philipp-hellwig/MLC-letter-string-analogies}.}

Our results show that these models can be trained to solve letter-string analogies using meta-learning (Table \ref{tbl:main_results}), performing on par with frontier LLMs, and generalizing very well to new alphabets and moderately well to new combinations.
However, we find that our models struggle to generalize to new transformations (Section \ref{sec:results}). In a series of detailed computational experiments we study the conditions for successful generalization to alternative alphabets. We show that including copy tasks in training facilitates this generalization by reducing shortcut learning, and also highlight the importance of enough variation in the alphabets during training. 

Finally, we look inside our successful  models, and perform a (mechanistic) interpretability analysis that gives  surprisingly detailed insights into \emph{how} the models perform their task.  We do not train LLMs, but we discuss how our understanding of how small transformers learn to solve letter-string analogies holds lessons for how to study and improve LLMs' analogical reasoning abilities.

\section{Background and Related Work}
\textbf{Analogy} has been a topic of research for decades \citep{gentner1983structuremapping,hofstadter1994copycat,hummel1997distributed}. Computational models of analogy have been developed for multiple domains, and approaches to modelling analogy include structure mapping \citep{gentner1983structuremapping}, the CopyCat project \citep{hofstadter1994copycat}, and methods that extract and construct relations \citep{doumas2008theory}. More recently, LLMs' ability to process analogy has been debated \citep{webb2023emergent,hodel2024response, stevenson2025can,lewis2024evaluating}, and approaches including
Meta-analogical Contrastive Learning \citep{kim2020fewshot}, model-based reinforcement learning \citep{lee2024enhancing}, program synthesis \citep{li2024combining}, 
 graph representations and semantic mapping theory have been proposed \citep{crouse2021neural, ling2022deepgar}.
 
\textbf{Letter-string analogies}, introduced by \citet{hofstadter1994copycat}, form an interesting task to study analogical reasoning in language models and humans, given that they are text-based, require little domain knowledge and can be used to study abstraction, generalization and compositionality in both humans and models \citep{lewis2024evaluating, hofstadter1983metamagical}. Letter-string analogies take place in the context of a given alphabet, and 
consist of an example that illustrates how a string of letters is transformed to a new string of letters (e.g., \textit{a b c d} $\rightarrow$ \textit{a b c e}), followed by a query (e.g. \textit{i j k l} $\rightarrow$  ?) where the agent must apply the transformation to the new string analogous to the example. For this particular task, the most commonly given answer is \textit{i j k m} (``replace the last letter with the next letter in the alphabet''); however, alternate rules also exist such as \textit{i j k e} (``replace the last letter with \textit{e}'').

Analogical reasoning and, specifically, the ability to solve letter string analogies has become one key issue in the debate about whether Transformer-based language models are exhibiting \textbf{`true' reasoning behavior} \citep[e.g.,][]{valmeekam_planning_2023,lewis2024evaluating,stevenson2025can,shojaee_illusion_2025,xu_are_2025}. In that debate, an important distinction must be made between the abilities of the core Transformers and those of complete agentic system, with access to symbolic tools and embedded in an agentic harness \citep{ke2025surveyfrontiersllmreasoning}. While recent advances with the latter are truly impressive, many open questions remain about how to precisely characterize the learned reasoning abilities of the Transformers at their core \citep[e.g.,][]{langedijk2025propositional,sun2025climbing}. In the current paper, we focus on a core (encoder-decoder) Transformer, and ask whether and how it may learn to represent analogical reasoning.

\textbf{Meta-Learning} is the process of training models to \textit{learn how to learn}. Given the breadth of the definition, meta-learning encompasses many different approaches \citep{hospedales2022metalearning}, such as 
memory augmented neural-networks \citep{santoro2016metalearning}, meta-reinforcement learning \citep{duan2016rl$^2$, liu2021decoupling}, and model-agnostic meta-learning (MAML) \citep{finn2017modelagnostic}. 
The meta-learning method we apply in this study is called Meta Learning for Compositionality\citep[MLC,][]{lake2023humanlike}. 
MLC works similarly to MAML and claims to generalize in a human-like and systematic fashion. 

\citet{lake2023humanlike} used MLC to train models on a grammar-learning task where rules had to be deduced to transform an input sequence of pseudowords to an output sequence of colored dots, given a set of examples (i.e., primitives and functions). 
For example, "Input: dax = \red{$\bullet$}
; lug = \green{$\bullet$}; wif = \blue{$\bullet$}; dax fep = \red{$\bullet\bullet\bullet$}; lug fep = \green{$\bullet\bullet\bullet$}; Query: wif fep ? Output: \blue{$\bullet\bullet\bullet$}".
The authors characterized this procedure as meta-learning because %
the model had to learn how to solve this \emph{type} of task, rather than learning only the specific task. 
\citet{lake2023humanlike} report performance of up to 100\% accuracy on this task. Additionally, the models also managed to solve novel functions that implemented previously unseen transformations. 

\textbf{Meta-Learning for Analogy.} We investigate whether using MLC helps transformers learn to solve letter-string analogies. %
We incorporate the meta-learning aspect by using different permuted alphabets \citep{lewis2024evaluating}.
However, an additional challenge for letter-string analogies compared to the \citet{lake2023humanlike} grammar-learning task, is that in our task the transformation cannot be deduced from the provided primitives and functions, but must be inferred from the example in the context of the given alphabet (i.e., in-context learning). Furthermore, many frontier models struggle with transformations that are easy for humans (e.g., Example: \textit{d f e g} $\rightarrow$ \textit{d e f g } Query: \textit{m l n o} $\rightarrow$ ? Answer: \textit{l m n o}) \citep{lewis2024evaluating}. Therefore, the question arises whether the MLC approach improves transformer performance on transformations that frontier models typically struggle with.

\section{Method}\label{sec:method}

\subsection{Architecture and Training}

We used the same model architecture (a standard sequence-to-sequence transformer) as in \citet{lake2023humanlike}. To adapt the model to letter-string analogies, the encoder’s vocabulary consists of the 26 letters of the English alphabet (\textit{a, b, c, d, $\dots$, z}), a padding token (PAD), a token to separate examples from each other ($|$), and a token ($\rightarrow$) to show how an input string of letters transforms to an output string of letters. The decoder’s alphabet is the same as the encoder's except for the tokens ($|$, $\rightarrow$) and instead, the decoder’s vocabulary will contain a start-of-sequence- and an end-of-sequence token ($\rightarrow$, EOS). 

\begin{figure}[!ht]
    \centering
    \includegraphics[width=\linewidth]{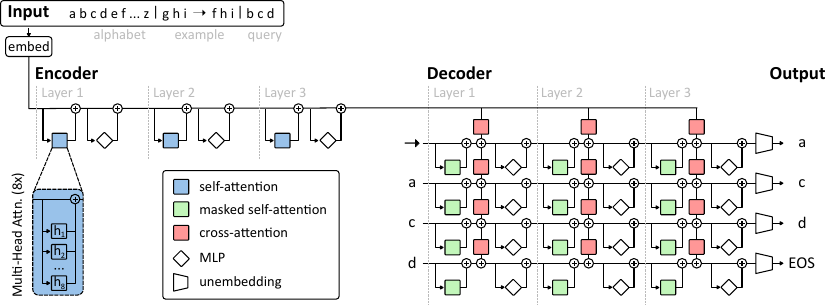}
    \caption{A forward pass through the model, with an example letter-string analogy task.}%
    \label{fig:arch}
\end{figure}

For an illustration of a forward pass through the MLC model, see Figure \ref{fig:arch}.
Each letter is represented by a 128-dimensional embedding. 
The encoder and decoder consist of 3 attention layers with 8 heads each. In between each multi-head attention block sits a 512-dimensional feed-forward layer. After the final layer of the decoder comes a final feed-forward layer that shrinks the dimensionality to the size of the vocabulary which is used as the logits in a softmax probability distribution used for next token predictions. All MLC models were implemented in PyTorch \citep{paszke2019pytorch}, using the Transformer class for the encoder and decoder.

We trained the models using a batch size of 32. Initial hyperparameter selection (learning rate, epochs, model size) had minimal impact on performance, therefore, to keep results comparable, we kept the training configuration as close as possible to the original MLC paper \citep{lake2023humanlike}. This includes: learning rate of 0.001 with linear decay applied after each epoch and dropout of 0.1. Training runs converged relatively quickly, so we shortened the training duration from 50 epochs used in \citet{lake2023humanlike} to 20 epochs.

\subsection{Data Generation}\label{sec:data_gen}

We generate a dataset of letter-string analogy problems split into training, validation, in-distribution test and out-of-distribution test. Each split comprises a number of transformations combined with a number of different alphabets. As illustrated in the example below, an individual task is composed of an alphabet, an example input-output pair and a query input where the model has to predict the correct continuation that completes the query analogically to the example.

$$ \underbrace{\textit{a b c d ... z }}_{\text{alphabet}} |\underbrace{\textit{ g h i} \rightarrow \textit{f h i }}_{\text{example}} | \underbrace{\textit{b c d}}_{\text{query}} \rightarrow \quad ? $$

\paragraph{Alphabets and Transformations}Since most transformations depend on which alphabet is used, we use multiple permuted \textbf{alphabets} to generate our datasets (based on \citet{lewis2024evaluating} with 0-, 2-, 5-, 10-, and 20 permutations to the standard alphabet). For our training split, we generated examples by sampling a substring of the alphabet (lengths 2 to 6) and applying a training transformation to it (shown in Table \ref{fig:transformations}, left). To generate the queries, we shuffled the examples and only retained query and example pairs that differed.

\begin{figure}
\resizebox{\textwidth}{!}{
\begin{minipage}[t]{0.57\textwidth}
\begin{tabular}[t]{p{3.3cm}l p{3cm}l}
\toprule
\multicolumn{2}{c}{\bf Seen Transformations in Training} \\
\textbf{Transformation}           &\textbf{Example}\\ \midrule
Extend (Ext)             &a b c $\rightarrow$ a b c d \\
Successor (Succ)         &a b c $\rightarrow$ a b d\\
Predecessor  (Pred)      &b c d $\rightarrow$ a c d\\
Remove Redundant (RmRed) &a b b c $\rightarrow$ a b c\\
Fix                      &a b w d $\rightarrow$ a b c d\\
Sort                     &a d c b $\rightarrow$ a b c d\\
Sort + Group (Gp)        &d d c c b b $\rightarrow$ b b c c d d\\
RmRed + Interleave (IL)  &a x b x b x c$\rightarrow$ a x b x c\\
RmRed + Succ             &a b b c $\rightarrow$ a b d\\
Fix + Ext                &a b w d $\rightarrow$ a b c d e \\
\bottomrule
\end{tabular}
\end{minipage}
\begin{minipage}[t]{0.57\textwidth}
\begin{tabular}[t]{p{2.7cm}l p{2.1cm}l}
\toprule
\multicolumn{2}{c}{\bf New Compositions of Seen Transformations} \\
\textbf{Transformation} & \textbf{Example} \\ \midrule
RmRed + Sort     &{a d d c b$\rightarrow$ a b c d}\\ %
Ext + Pred       &{b c d$\rightarrow$ a c d e}\\%Extend Sequence + Predecessor\\
Fix + IL         &{a f b f w $\rightarrow$ a f b f c}  \\ %
Ext + Gp         &{a a b b$\rightarrow$ a a b b c c}\\% ExtSeq + group
Ext + Ext + Succ &{a b c$\rightarrow$ a b c d f}\\ %
Fix + Pred + Succ&{b c w e $\rightarrow$ a c d f}\\ %
\midrule
\multicolumn{2}{c}{\bf Novel Transformations} \\
\textbf{Transformation}& \textbf{Example} \\ \midrule
Reverse (Rev)    & {a b c $\rightarrow$ c b a} \\
Shift            &{a b c$\rightarrow$ d e f} \\
Replicate (Rep)  & {a b c $\rightarrow$ a b c a b c}\\
\bottomrule
\end{tabular}
\end{minipage}
}
    \captionof{table}{The full list of transformations used in this study. Transformation examples are shown for the standard alphabet only. `+' is used as a shorthand to describe \emph{successive} application of multiple transformations. The left side contains transformations that the models are trained on; the right column contains transformations that models hadn't seen during training (compositional and novel).}
    \label{fig:transformations}
\end{figure}

This procedure generated 459,013 tasks per dataset, 367,211 of which were randomly assigned to the training set, 45,901 to the validation and \textbf{in-distribution test set} respectively. Along with our in-distribution test, we also evaluate how well models generalize to \textbf{out-of-distribution tests}: (1) how well they generalize to problems that are based on new alphabets, not contained in the training data, and 
(2) whether the trained models can solve new transformations: we test them on 6 combinations of seen transformations (compositional transformations) and 3 unseen transformations (novel transformations), listed in Table \ref{fig:transformations}, right.

\subsubsection{Dataset Variants}

We create dataset variants along the three different factors described below, an overview of the dataset variants is shown in Table \ref{tab:dataset_variants}.

\textbf{Copy Tasks}
Inspired by \citet{lake2023humanlike}, we study the effects of including copy tasks in the training data, where the example and the query are the same. %
In this alternative dataset, half of the tasks are copy tasks, while the example input and query input differ for the other half. 

\textbf{Number of Seen Alphabets} To evaluate the effect of alphabet diversity, we created dataset variants ranging from 20 to 400 seen permuted alphabets. To isolate the effect of increasing the number of permuted alphabets, we held the overall dataset size constant.

\textbf{Few-shot Learning} To check how and whether the number of examples affects performance, we gradually increased the number of examples the model was given for each task from 1 to 5. However, increasing the number of examples resulted in worse performance of our models (see Appendix \ref{app:fewshottraining} for details). We also addressed whether few-shot learning improved LLM performance on letter-string analogies. The results were mixed, but the best performing LLM, GPT-OSS 120B also performed worse with more in-context examples (see Appendix \ref{app:llmXtransformation}). 

\begin{table}[htbp]
  \centering
  \begin{tabular}{ccc}
    \toprule
    \textbf{Copy Tasks} & \textbf{Num. Seen Alphabets} & \textbf{Num. Examples} \\
    \midrule
    \color{red}{\xmark}  & 20  & 1\\
    \color{Green}{\cmark} & 20  & 1\\
    \color{Green}{\cmark} & 40  & 1\\
    \color{Green}{\cmark} & 60  & 1\\
    \color{Green}{\cmark} & 80  & 1\\
    \color{Green}{\cmark} & 100 & 1\\
    \color{Green}{\cmark} & 140 & 1\\
    \color{Green}{\cmark} & 200 & 1\\
    \color{Green}{\cmark} & 400 & 1\\
    \color{Green}{\cmark} & 20  & 2\\
    \color{Green}{\cmark} & 20  & 3\\
    \color{Green}{\cmark} & 20  & 4\\
    \color{Green}{\cmark} & 20  & 5\\
    \bottomrule
  \end{tabular}
  \caption{Overview of all the dataset variants used for training.}
  \label{tab:dataset_variants}
\end{table}

\section{Behavioral Results}\label{sec:results}

For all of the following results, we evaluated the MLC models with frozen weights. We obtained predictions from the models by sampling the token with the largest probability at each prediction step until the EOS token was sampled or until the response length exceeded the longest target in the batch.

\begin{figure}[hb]
  \centering
  \begin{minipage}[t]{0.60\textwidth}
    \centering
    \vspace{0pt}
    \begin{tabular}{llllll}
        \toprule
        copy & number & \multicolumn{2}{c}{seen transform} & \multicolumn{2}{c}{new transform} \\
        tasks & seen $\alpha$'s  & seen $\alpha$ & new $\alpha$ & seen $\alpha$ & new $\alpha$ \\
        \midrule
        \color{red}{\xmark} & 20 & 60.9 & 26.5 & 10.2 & 3.2 \\
        \color{Green}{\cmark} & 20 & 91.0 & 33.1 & 10.2 & 3.9 \\
        \color{Green}{\cmark} & 200 & \textbf{96.7} & \textbf{93.9} & \textbf{13.1} & \textbf{12.3} \\
        \bottomrule
    \end{tabular}
    \captionof{table}{Performance comparison on in- and out-of distribution test sets. Results are split based on differing factors during training. Accuracies are averaged over 10 replication runs (highest averages in bold). Full set of results, including additional variants and 95\% bootstrapped confidence intervals in Appendix \ref{apx:detailed_training}. $\alpha$=alphabet. }
    \label{tbl:main_results}
  \end{minipage}
  \hfill
  \begin{minipage}[t]{0.38\textwidth}
    \centering
    \vspace{0pt}
    \includegraphics[width=\textwidth]{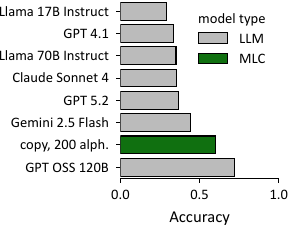}
    \captionof{figure}{Our best model (green) on par with LLMs (grey) on our aggregated dataset (Table \ref{tbl:main_results}).}
    \label{fig:llm_comparison}
  \end{minipage}
\end{figure}

\paragraph{Including copy tasks facilitates learning.}
We find that including \textbf{copy tasks} in the dataset gives the models a considerable boost in performance on seen alphabets and seen transformations (Table \ref{tbl:main_results} first row compared to second row). 
However, performance still declined considerably for tasks that were generated from new permuted alphabets (from 90.7 to 33.3\%) and accuracy on new transformations stayed fairly low in general. Overall, the MLC models that were trained using copy tasks still generalized poorly to new contexts. %

\begin{wrapfigure}[17]{r}{0.45\textwidth} %
    \centering
    \includegraphics{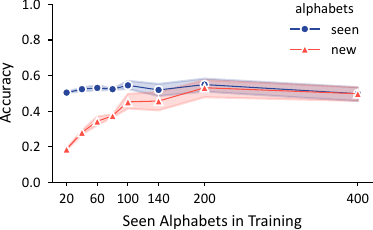}
    \caption{Increasing the number of seen alphabets in training leads to better generalization to new alphabets (red and blue lines converge). Bands indicate 95\% bootstrapped confidence intervals with 10 replications.}
    \label{fig:num_seen_alphabets}
\end{wrapfigure}

\paragraph{Models can generalize to novel 26-letter alphabets, with sufficient variation at training time.} 
Our results indicate that increasing the number of alphabets in training improves generalization to new alphabets and new transformations: Figure \ref{fig:num_seen_alphabets} shows that when the number of alphabets seen in training are increased, accuracies on seen- and new alphabets converge to similarly high accuracies. Furthermore, our best training run (200 alphabets and copy tasks) specialized for this task outperforms most general-purpose frontier models on our dataset (Figure \ref{fig:llm_comparison}).

\paragraph{Models struggle with generalizing to new transformations.}

\begin{figure}[h]
    \centering
    \includegraphics[width=\linewidth]{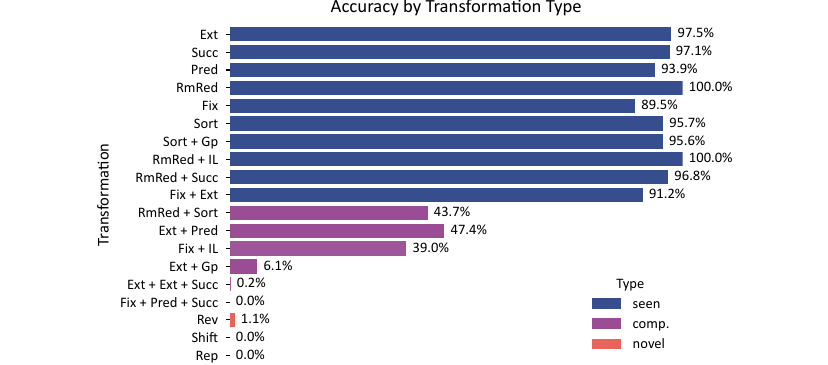}
    \caption{Averaged accuracies split by transformation with the best performing models with 10 replication runs (trained on a dataset with 200 permuted alphabets with copy tasks).}%
    \label{fig:transform_acc}
\end{figure}

Figure \ref{fig:transform_acc} shows how well the models trained on a dataset with 200 permuted alphabets perform on each transformation. Even though the compositional transformations are not specifically optimized for, the model learned to correctly solve some of the transformations composed of two seen transformations. However, transformations composed of three seen transformations and novel transformations were not learned. Our model generally outperforms tested frontier models on compositional transformations (see Appendix \ref{app:llmXtransformation}).

\section{Interpretability Analyses}\label{sec:analyses}

\begin{figure}[t]
   \centering
   \includegraphics[width=\linewidth]{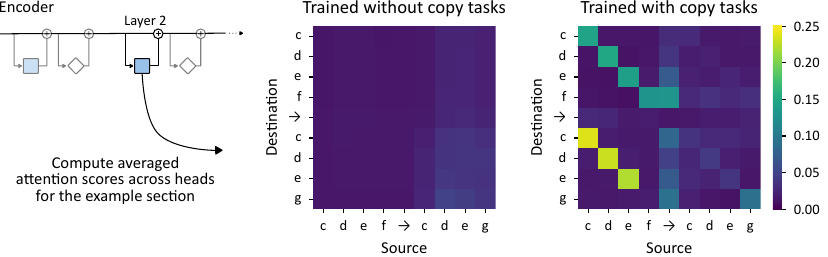}
   \caption{Models trained without copy tasks tend to show no information flow between example input and output (i.e., near-zero attention weights); models trained with copy tasks do.}
   \label{fig:enc_attn}
\end{figure}

\paragraph{The effect of including copy tasks}
{Why do we see a large performance boost when training with copy tasks?} A key difference between training with and without copy tasks is the formation of \emph{matching heads} (reminiscent of the \emph{induction heads} in LLMs; \citealp{olsson2022context}). Figure \ref{fig:enc_attn} shows the average attention scores of the heads of layer 2 in the encoder. Here, we see low attention scores between tokens of the example input-output pairs when training without copy tasks for most replication runs (center, see Appendix \ref{app:matching_head_occurrence} for more details). In contrast, we observe a pattern of high attention scores between matching example input-output pairs %
when copy tasks are included for all replication runs (right).

\paragraph{Obtaining a mechanistic understanding of the induced algorithm}  %

How do the trained transformers mechanistically solve the tasks? To better understand the \emph{induced} algorithm, it is useful to first briefly consider a simple, symbolic algorithm for the \emph{predecessor} task:
\begin{list}{}{}
    \item[0] \textbf{(Initialization)} Input: string \texttt{s = alphabet|ex\_in>ex\_out|query} \\Example: \texttt{abcdefghij|ghi>fhi|bcd}
    \item \begin{list}{}{}
        \item[1] \textbf{(Matching)} Get indices of \texttt{ex\_in}, \texttt{ex\_out}, and \texttt{query} in \texttt{alphabet}. 
        \item Call these \texttt{ex\_in$_\texttt{inds}$}, \texttt{ex\_out$_\texttt{inds}$}, and \texttt{query$_\texttt{inds}$}\\Example: \texttt{ghi}$\mapsto$\texttt{[6, 7, 8]}; \texttt{fhi}$\mapsto$\texttt{[5, 7, 8]}; \texttt{bcd}$\mapsto$\texttt{[1, 2, 3]};
        \item[2] \textbf{(Compute transformation)} Compute: \\ \texttt{ans$_\texttt{inds}$}=\texttt{concat(}[\texttt{ex\_out$_\texttt{inds}$[0]}-\texttt{ex\_in$_\texttt{inds}$[0]}+\texttt{query$_\texttt{inds}$[0]}], \texttt{query$_\texttt{inds}$[1:])}\\
        Example: \texttt{ans$_\texttt{inds}$} = \texttt{concat([5 - 6 + 1], [ 2, 3]} = \texttt{[0, 2, 3]}
        \item[3] \textbf{(Apply transformation)} Index into \texttt{alphabet} with \texttt{ans$_\texttt{inds}$} to obtain \texttt{ans}. \\
        Example: \texttt{alphabet[0, 2, 3]$\mapsto$acd}
    \end{list}
\end{list}

By implementation in RASP-L \citep{zhou2024rasp}, we find that it is possible for a transformer to carry out this algorithm (Appendix \ref{apx:raspl_impl}). We can then ask: %
does our trained model compute \textit{predecessor} this way? %

We studied the trained transformers using a variety of methods, including attention visualization and interchange interventions. We found that the induced algorithm can be understood in great detail. Here we discuss evidence for equivalents of each of the 4 steps above, continuing with the predecessor transformation as a 
running example, %
with the standard alphabet, the example \textit{g h i $\rightarrow$ f h i} and query \textit{b c d $\rightarrow$}, which the model must complete with \textit{a c d}. %

\begin{figure}
    \centering
    \includegraphics[width=\linewidth]{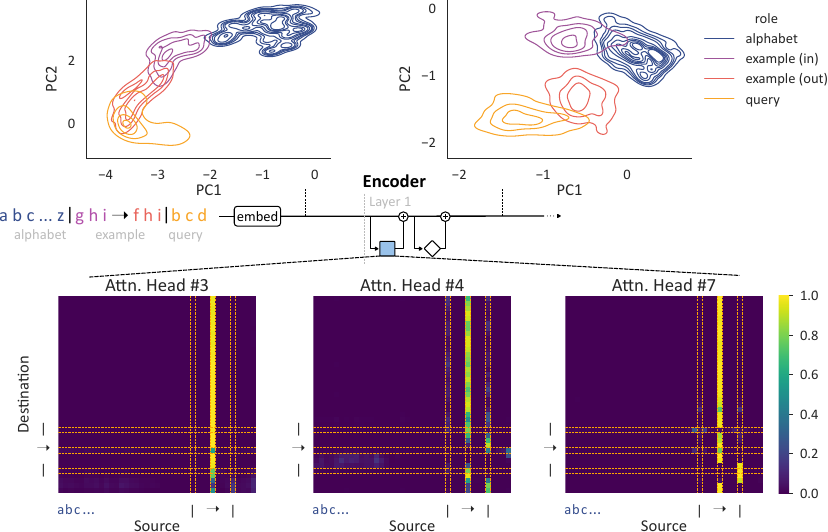}
    \caption{Illustration of the initialization step (where \emph{role assignment} happens) in the model. The upper part shows that the representations separate more cleanly after the first layer of the encoder. This is likely induced by the attention heads in the lower part of the figure.}
    \label{fig:role_assignment}
\end{figure}

\paragraph{Initialization}
We identify multiple heads in layer 1 of the encoder that have characteristic patterns of attention. %
In particular, the separator tokens ($|$, $\rightarrow$) show high attention weights with large sections of the input, which results in vertical stripes (see Figure \ref{fig:role_assignment}, lower part). We hypothesize that the attention heads may implement \emph{role assignment} in this phase through position comparisons. Since we use positional encoding in our transformers, the position of each token in the sequence is present in their vector representations. When a token attends to a separator, part of the vector representation of the separator (relative to the $V$ matrix and the attention score $a$) is added to its representation. The MLP component could then compare the parts of the vector representations that correspond to the position of the token and the separator and identify whether this token comes before or after it in the sequence.

If this process really occurs in these heads, we would expect that after these heads have been processed, token representations separate more cleanly based on their role. We test this hypothesis by training a principal component analysis (PCA) with two components on residual stream activations across all layers of the encoder. Afterwards, we apply this PCA to the vector representations on a test set of tasks before and after layer 1 of the encoder. In Figure \ref{fig:role_assignment}, we group tokens according to their roles and visualize them using kernel density estimates. We see that the alphabet, example-in, example-out and query already cluster to some degree before the first layer but cluster much more clearly after the first layer (in particular, the example-out no longer overlaps with the example-in and the query).
    
\paragraph{Matching}
\begin{figure}[!ht]
    \centering
    \includegraphics[width=\linewidth]{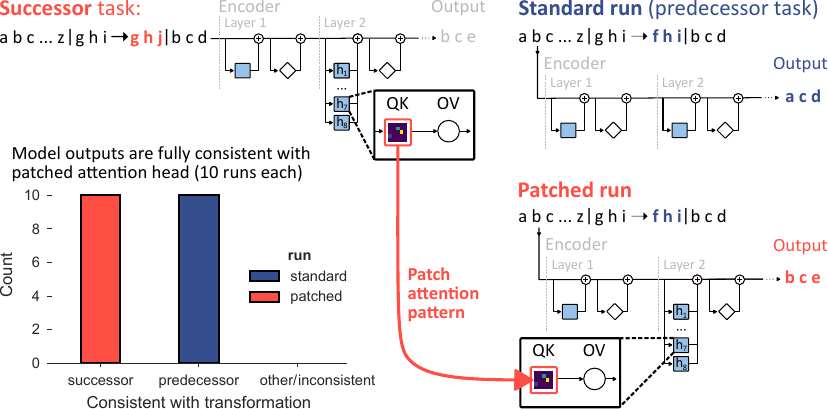}
    \caption{Results from our \textbf{attention pattern patching}. %
    A single attention head in the second encoder layer implements the \emph{matching} operation of the algorithm. We demonstrate the effects of this head %
    are \emph{causal}, by replacing the standard attention matrix $A^{\text{pred}}$ with $A^{\text{succ}}$ produced by a successor task.}
    \label{fig:matching_intervention}
\end{figure}

We find that %
a single attention head \#7 in layer 2 of the encoder implements the \emph{matching} operation: the same letters match with each other if they have different roles, otherwise they match with themselves leading to high attention scores in these token-pairs. 
We verify the causal role of this head %
in the outputs of the model using \textbf{attention pattern patching} (using the package \emph{NNSight} \citep{fiotto-kaufman2025nnsight}), as detailed in Figure~\ref{fig:matching_intervention}.

\begin{figure}[!ht]
    \centering
    \includegraphics[width=0.9\linewidth]{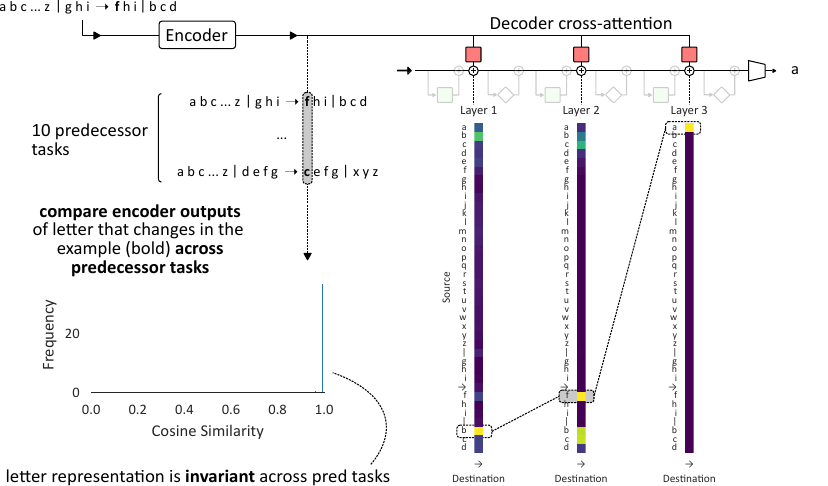}
    \caption{The \emph{Compute mapping and Apply mapping} steps of the algorithm. The encoder builds an invariant representation for the predecessor transformation and stores it in the first letter of the example output (left). This representation is then used in the decoder cross-attention (layer 2) to move attention from letter 'b' (layer 1) to letter 'a' (layer 3) of the decoder.}
    \label{fig:mapping}
\end{figure}

\paragraph{Compute mapping and Apply mapping}
The \emph{Compute mapping} step is implemented in the 3rd layer of the decoder, inside the residual stream for the tokens of the example-output. The  \emph{Apply mapping} step happens in the decoder cross-attention. The decoder is initialized with the token $\rightarrow$; in the first layer, this token attends to the first token of the query (\textit{b}). In the second layer, it attends to the first token of the example-output (\textit{f}). The process is depicted in Figure~\ref{fig:mapping}.

To test whether our analysis is correct, we carried out one more diagnostic experiment. We investigated whether the residual stream representations in the encoder's 3rd layer for the example-out tokens are invariant to the identity of the first token of the example output (i.e., the letter \textit{f} in our running example) across predecessor tasks, as these must be invariant if they are to represent the computed mapping. This is indeed the case, as evidenced by cosine similarity close to 1 for all comparisons (Figure \ref{fig:mapping}, bottom-left).

\section{Discussion \& Conclusions}
We examined whether meta-learning for compositionality \citep[MLC,][]{lake2023humanlike} can help small transformer models solve letter-string analogies, finding that our MLC trained model can (i) reliably solve trained analogy transformations, (ii) generalize robustly to novel 26-letter alphabets when trained on sufficiently heterogeneous data, and (iii) generalize to some compositions of trained transformations, but (iv) still fails on entirely novel transformations. Mechanistic interpretability analyses show that our models implement a relatively simple, algorithmic procedure for solving the analogies. 

Our results complement work showing that transformer-based LLMs can solve letter-string analogies, but in a highly context-dependent way. However, where LLMs struggle with novel alphabets and complex compositions of transformations \citep[e.g.,][]{webb2023emergent, hodel2024response, lewis2024evaluating, stevenson2025can}, our work shows that the MLC approach mitigates some of these shortcomings, where our best model performs on par with frontier LLMs, including LLMs in multishot settings. Our evaluation was limited to generalization scenarios close to the original task format. Future work could examine how well our models generalize to different alphabet lengths or novel symbols.

Mechanistically, our models implement an algorithm akin to word2vec-style vector arithmetic, also observed in LM analogical reasoning \citep{Merullo2023analogyvectorarithmetic}. This provides a complementary, more controlled demonstration that transformer architectures can internalize simple, interpretable relational algorithms.

Classical theories in cognitive science characterize analogical reasoning as involving two key processes: forming abstractions and mapping relations \citep{gentner1983structuremapping, gentner2017analogy, petersen2025modellinganalogycogscinlp}. Our models instantiate computational analogues of these steps: the learned algorithm operates over abstract relational transformations defined over string positions rather than specific letters, paralleling the abstraction step in which reasoners construct relational schemas that transcend instances \citep{gentner2017analogy}. The model then maps example input–output pairs to the query by aligning structural roles (e.g., “first letter”, “successor”) and transferring the inferred transformation, mirroring Gentner’s notion of mapping from base to target while preserving relational structure \citep{falkenhainer1989structuremappingengine}.

Despite focusing on small models and toy analogies, we believe our findings are also relevant for larger models. In particular, our insights into copy tasks and dataset heterogeneity could inform the design of pretraining curricula for larger models. For example, combining next-token prediction with meta-learning curricula that stress relational abstraction and mapping over varied symbol systems during pretraining could help larger models acquire more human-like analogical reasoning capabilities.

\section*{Acknowledgements}

We would like to thank Gustaw Opiełka for his helpful comments on the analysis section of our paper as well as Michael Hanna for his advice on the implementation of the interpretability analyses. ML thanks Patrick Gemmell for extensive discussions on RASP.

\bibliography{references}
\bibliographystyle{colm2026_conference}

\appendix

\section{Detailed Results on Training Conditions}
\label{apx:detailed_training}
\textbf{Batching Methods}
In addition to the training conditions mentioned in the main text, we also experimented with different batching strategies during training. These included \textit{random} batching, where tasks were drawn randomly to construct a batch (default condition), batching by \textit{alphabet}, where tasks were drawn randomly with the alphabet held constant for each task in the batch, and some other variations. Batching methods had a negligible impact (see \ref{apx:detailed_training}), therefore we report aggregated results over multiple batching methods.
In Table \ref{tbl:detailed_main_results}, we present a more detailed overview of our training experiments. For our initial training regimes (20 training alphabets with- and without copy tasks), we used four different batching strategies. When we gradually increased the number of training alphabets, we only retained \textit{random} batching as a baseline and batching by \textit{alphabet} since it gave the better results in early experiments. For each level, we conducted 5 replication runs. 

Batching by alphabet with 200 permuted alphabets in the training set gave the best results in terms of generalization to new transformations while still showing good performance on seen transformations (Table \ref{tbl:detailed_main_results}, last row).  \textit{Random} batching had slightly better performance on seen transformations, performance was lower for new transformations across batching methods.

\begin{table*}[ht]
    \centering
    \rotatebox{90}{
    \begin{tabular}{lllllll}
        \toprule
        copy & num. seen & \multirow{2}{*}{batching method} & \multicolumn{2}{c}{accuracy seen transform. (\%)} & \multicolumn{2}{c}{accuracy new transform. (\%)} \\
        tasks & alphabets &  & seen alphabets & new alphabets & seen alphabets & new alphabets \\
        \midrule
        \multirow{4}{*}{\color{red}{\xmark}} & \multirow{4}{*}{20} & alphabet & 56.3 \small{[55.5, 57.5]} & 27.8 \small{[27.8, 27.9]} & 10.2 \small{[9.6, 10.9]} & 3.6 \small{[3.2, 4.1]} \\
         &  & random & 65.3 \small{[55.9, 79.4]} & 27.7 \small{[26.3, 29.9]} & 9.4 \small{[6.9, 12.5]} & 3.2 \small{[2.3, 4.2]} \\
         &  & transform. & 64.5 \small{[63.1, 65.6]} & 23.3 \small{[23.0, 23.8]} & 13.6 \small{[11.9, 15.7]} & 3.2 \small{[2.8, 3.6]} \\
         &  & transform. \& alph. & 57.6 \small{[55.2, 59.8]} & 27.2 \small{[26.5, 27.9]} & 7.6 \small{[6.0, 9.3]} & 2.7 \small{[2.1, 3.5]} \\
        \midrule
        \multirow{4}{*}{\color{Green}{\cmark}} & \multirow{4}{*}{20} & alphabet & 90.2 \small{[89.7, 90.7]} & 32.4 \small{[31.2, 33.3]} & 10.0 \small{[9.6, 10.5]} & 3.8 \small{[3.5, 4.1]} \\
         &  & random & 91.5 \small{[90.4, 92.6]} & 33.7 \small{[32.6, 34.8]} & 10.1 \small{[9.9, 10.2]} & 3.9 \small{[3.8, 4.0]} \\
         &  & transform. & 92.6 \small{[88.9, 96.0]} & 31.4 \small{[27.2, 35.0]} & 10.7 \small{[9.7, 11.8]} & 3.9 \small{[3.3, 4.9]} \\
         &  & transform. \& alph. & 89.2 \small{[88.2, 89.8]} & 34.3 \small{[32.0, 36.3]} & 10.1 \small{[9.2, 11.4]} & 4.0 \small{[3.4, 4.8]} \\
        \midrule
         \multirow{2}{*}{\color{Green}{\cmark}} & \multirow{2}{*}{200} & random & \textbf{97.7} \small{[93.7, 99.9]} & \textbf{94.5} \small{[86.3, 99.4]} & 10.4 \small{[8.6, 12.0]} & 9.9 \small{[7.8, 11.6]} \\
         &  & alphabet & 95.8 \small{[87.3, 100.0]} & 93.4 \small{[80.6, 99.9]} & \textbf{15.8} \small{[8.7, 22.6]} & \textbf{14.8} \small{[8.1, 21.6]} \\
        \bottomrule
    \end{tabular}
    }
    \caption{Comparison of different batching methods, including copy tasks in the training set, and increasing the number of seen alphabets during training. Accuracies are averaged over five replication runs (highest averages per column in bold). Numbers in square brackets signify 95\% bootstrapped confidence intervals.}
    \label{tbl:detailed_main_results}
\end{table*}
\clearpage
\section{Effect of Copy Tasks on Matching Head Development}
\label{app:matching_head_occurrence}
\begin{figure}[!ht]
    \centering
    \includegraphics[width=0.6\linewidth]{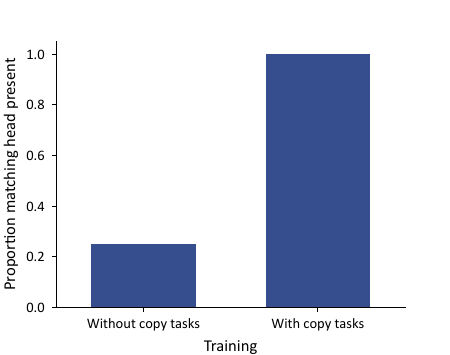}
    \caption{Proportion of runs that develop a matching head split by training with and without copy tasks (datasets with 20 seen alphabets).}
    \label{fig:matching_head_prop}
\end{figure}

To gauge the consistency of the effect of including copy tasks in training on the development of matching heads (presented in Figure \ref{fig:enc_attn}), we analyzed all 20 replications for the training datasets with 20 seen alphabets (5 replications for each of the 4 batching methods). Using a few random predecessor tasks, we visually inspected whether the attention patterns characteristic to the matching heads had occurred for each run. We scored relatively liberally so that whenever an attention pattern between the example inputs and outputs was present, we graded the model as has having developed a matching head. Overall, the weights and the location of the attention patterns varied moderately, sometimes occurring in multiple heads usually within the first two layers of the encoder. Figure \ref{fig:matching_head_prop} shows that matching heads rarely developed when training without copy tasks (25\%), but always developed when training with copy tasks (100\%).

\section{Failure Modes When Training Without Copy Tasks}
\label{app:failurenocopytasks}
\begin{table}[!ht]
    \begin{tabular}{llr}
        \toprule
        Transformation & Outcome & Prop. (\%) \\
        \midrule
        \multirow[t]{3}{*}{predecessor} & applied succ & 76.4 \\
        & correct & 14.9 \\
        & other incorrect & 8.7 \\
        \midrule
        \multirow[t]{2}{*}{successor} & applied pred & 85.1 \\
         & correct & 14.9 \\
        \bottomrule
    \end{tabular}
    \caption{Performance of one of the models trained without copy tasks on predecessor- and successor tasks. For almost all errors, the model applies the other transformation, suggesting that the model cannot distinguish between the two transformations.}
    \label{tbl:alt_rule_errors}
\end{table}

The fact that the models trained without copy tasks do not use the information contained in the example is further evidenced by their poor performance on predecessor- and successor tasks. For example, the model that used random batching had an accuracy of 14.9\% accuracy on successor- and predecessor tasks (see Table \ref{tbl:alt_rule_errors}). Most of the errors on predecessor tasks arise because the model applied the successor rule instead. Similarly, for all of the errors on predecessor tasks, the model applied the predecessor transformation. This likely occurs because the query inputs of both of these transformations are sorted subsequences of the alphabet (e.g., \textit{\textbf{c d e} $\rightarrow$ c d f} and \textit{\textbf{c d e} $\rightarrow$ b d e} respectively) and the query input length matches the query output length. If the model only uses the information contained in the query input to make its predictions, misclassifying predecessor transformations as successor transformations and vice versa is a logical consequence.
\section{Study of Attention Patterns}
\label{app:studyattention}
\begin{figure}[!ht]
    \centering
    \includegraphics[width=0.9\linewidth]{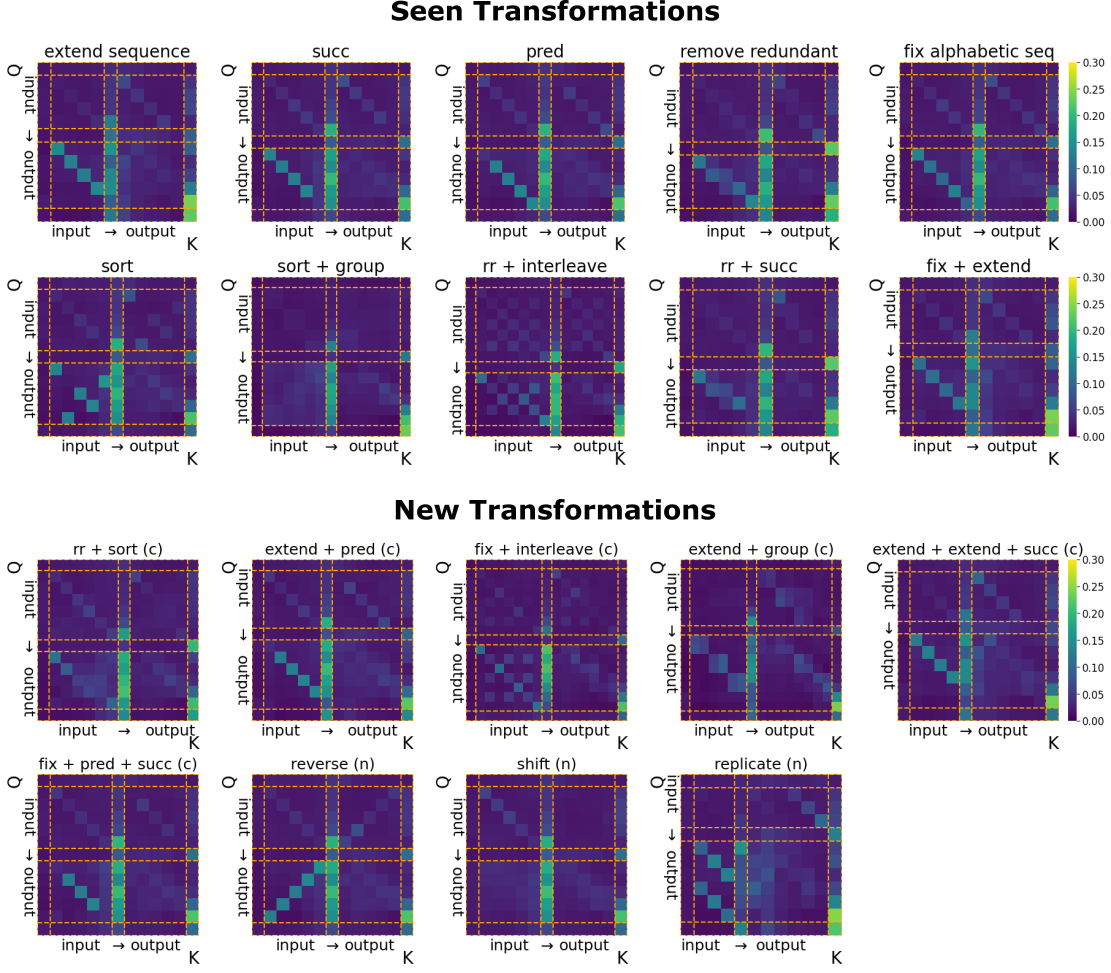}
    \caption{The best performing model forms distinct attention patterns for the example of each transformation type in the second encoder layer. For new transformations, (c) denotes compositional ones and (n) denotes novel ones. Where applicable, positions at which the transformation was applied were held constant across tasks of the same transformation.}
    \label{fig:study_attention}
\end{figure}

We found similar attention patterns when analyzing the best MLC model trained with tasks from 200 permuted alphabets. Figure \ref{fig:study_attention} shows the attention matrices for the examples in the second encoder layer. Each transformation is plotted separately, averaged over 20 tasks. Letters of the example output strongly attend to the same letter in the example input. For instance, the predecessor subplot ("pred") contains 5 letters in the input sequence as well as 5 letters in the output sequence (e.g., \textit{f g h i j $\rightarrow$ e g h i j}). In the bottom left quadrant, attention values are large for all elements of the input sequence except the first one, since it changes, for example, from \textit{f} in the example input to \textit{e} in the example output. 

Even for new transformations where the model performs poorly, the model shows characteristic attention patterns for each transformation (Figure \ref{fig:study_attention}. One example is the reverse transformation (e.g., \textit{a b c d e $\rightarrow$ e d c b a}). The last letter of the example output attends to the first letter of the example input and so on (resulting in an 'X' pattern). This suggests that these patterns alone are insufficient for good performance. 

\clearpage
\section{Few-shot Training Results}
\label{app:fewshottraining}
\begin{figure}[!ht]
    \centering
    \includegraphics[width=\linewidth]{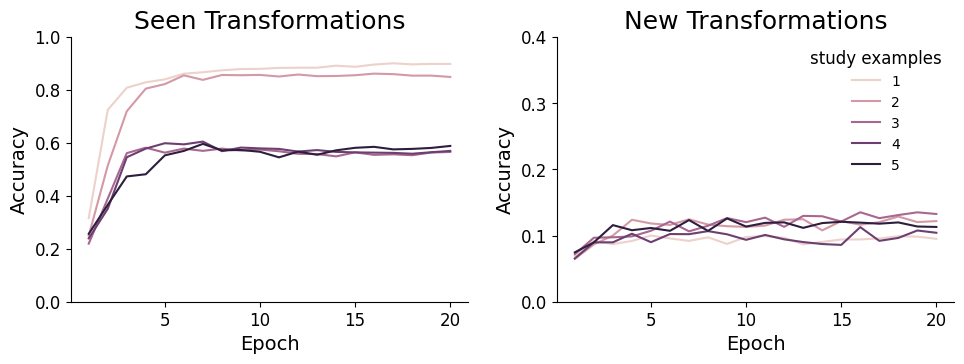}
    \caption{Validation accuracy declines when providing the model with more examples to solve each task.}
    \label{fig:study_examples}
\end{figure}

We trained models with increasing numbers of examples (from 1 to 5) on the dataset with 20 permuted alphabets (copy tasks included). Figure \ref{fig:study_examples} and Table \ref{tbl:study_examples} show that increasing the number of examples worsens the overall accuracy of 62.5\% when trained with one example per task to an overall accuracy of 37.1\% when trained with 5 examples per task.

\begin{table}[!ht]
    \centering
    \begin{tabular}{cr}
        \toprule
        examples & Accuracy (\%) \\
        \midrule
        1 & 62.5 \\
        2 & 53.7 \\
        3 & 36.2 \\
        4 & 35.2 \\
        5 & 37.1 \\
        \bottomrule
    \end{tabular}
    \caption{Overall accuracies of models trained with increasing number of examples per task.}
    \label{tbl:study_examples}
\end{table}

\subsection{Additional examples likely interfere with matching head formation}
We observed that the degradation is mainly tied to presence/absence of matching heads. Matching heads were generally present for models trained with 1 and 2 study examples where overall performance was relatively high (62.5 and 53.7\% accuracies). In contrast, we did not observe matching heads for models trained with 3, 4, and 5 examples which performed considerably worse (accuracies between 35.2 and 37.1\%). 

We believe the most plausible explanation for the performance degradation is that more examples make the auxiliary copy task harder. While the model can simply attend to the example when only one is provided, the model must additionally select the example where example input and query are the same. This added difficulty may interfere with matching head formation during training which likely explains the lower performance for models trained with more examples.

\clearpage
\section{Elimination Step \& Human Comparison}
\label{app:eliminationstep}

\begin{figure}[!ht]
    \centering
    \includegraphics[width=\linewidth]{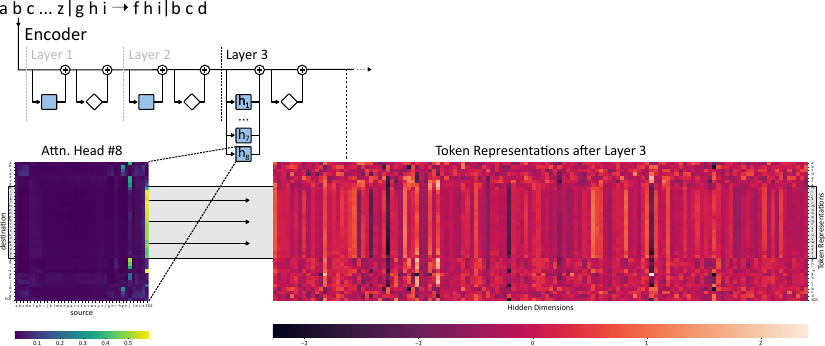}
    \caption{\emph{Elimination} step of the model where token representations that attend to the EOS token (left) are subsequently wiped out (right).}
    \label{fig:elimination}
\end{figure}

\paragraph{The model eliminates token representations of letters that are not required for the mapping step} Beside the three steps we discuss in the paper (role assignment, matching, and compute \& apply mapping), we also observed an elimination step that occurs in the final layer of the encoder. As illustrated in Figure \ref{fig:elimination}, representations of letters that attend strongly to the \textit{EOS} token (left) are wiped out in the outputs of the layer (right).

\begin{figure}[!ht]
    \centering
    \includegraphics[width=\linewidth]{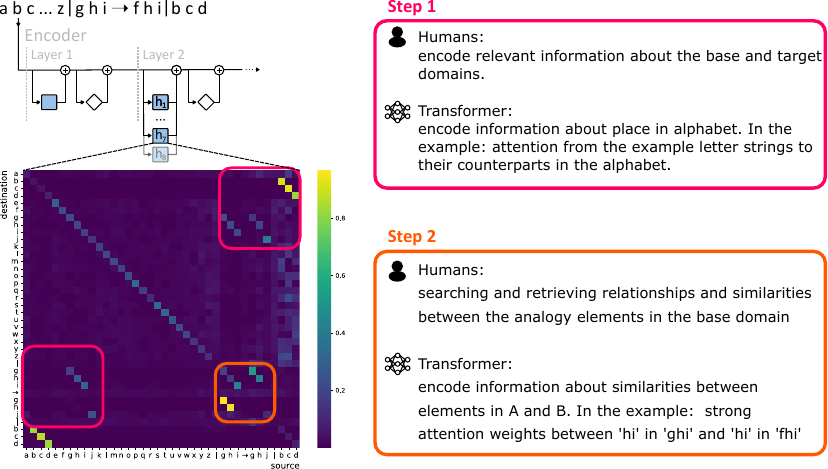}
    \caption{Comparison between \emph{matching step} and human steps in analogical reasoning.}
    \label{fig:human_similarities}
\end{figure}

\paragraph{Analogical reasoning steps of MLC models and humans overlap} We identify parallels between the matching step of the models from Section \ref{sec:analyses} and human analogy solving. Based on \citet{johnson2025large}, we describe the similarities to human analogical reasoning steps in Figure \ref{fig:human_similarities}. In particular, find that (1) encoding relevant information of the domains and (2) searching for and retrieving relationships and similarities between elements closely align with the attention patterns from head \#7 in the second layer of the encoder (Figure \ref{fig:human_similarities}, left).

\clearpage
\section{LLM versus MLC Models by Transformation Type}
\begin{figure}[ht]
    \centering
    \includegraphics[width=\textwidth]{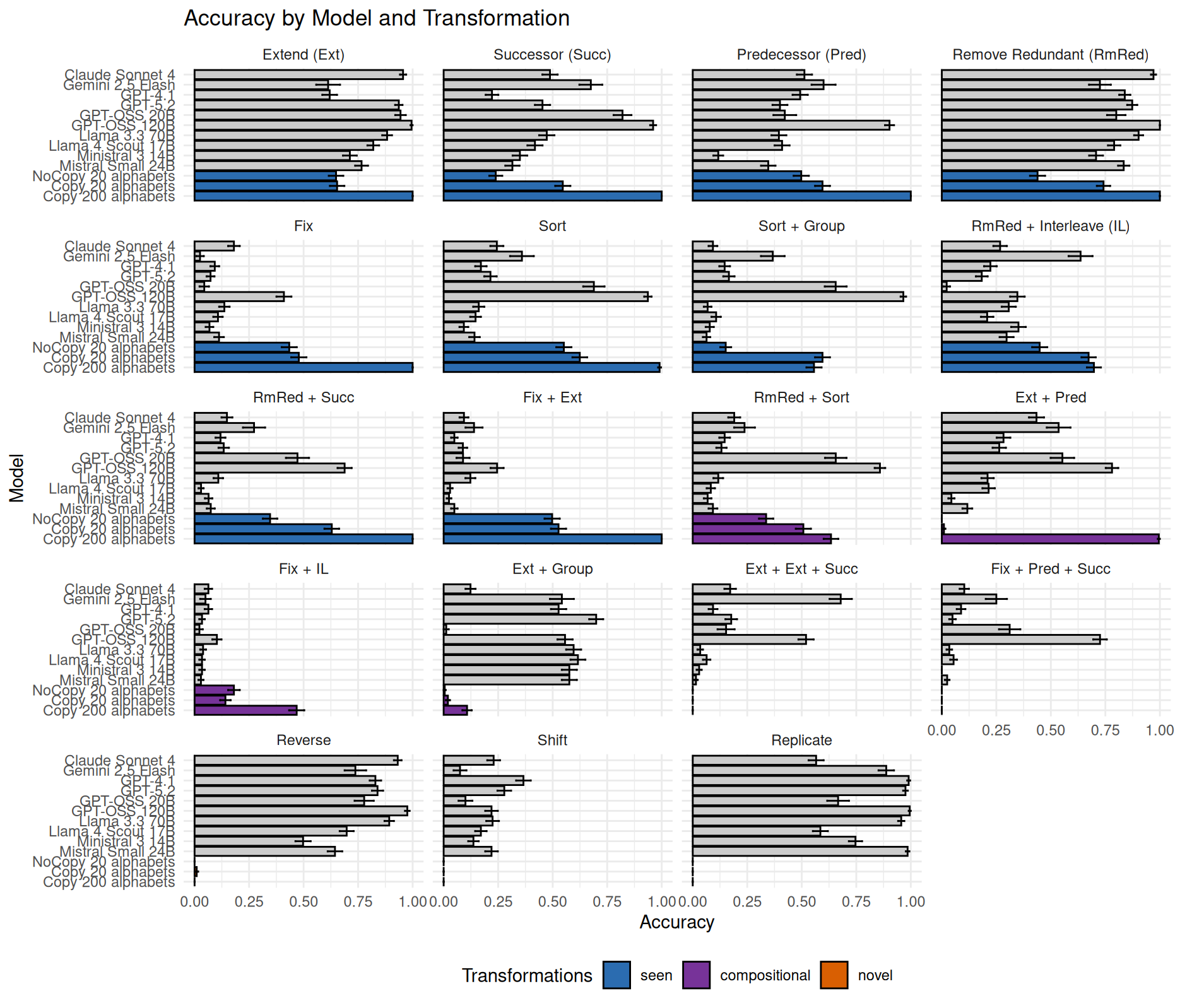}
    \caption{Frontier LLM  performance by Transformation Type compared to our MLC trained models.}
    \label{fig:acc_modelxtransformation}
\end{figure}

As can be seen in Figure \ref{fig:acc_modelxtransformation} our best MLC model, Copy 200 - trained with copy tasks on 200 alphabets, performs on par with frontier models on trained (compositional) transformations (blue coloured bars). For three of the four unseen compositional transformations with two trained transformations (purple coloured bars with 2 transformations) our model performs at or above the level of the best-performing frontier models (GPT-OSS models). However, for the "Extend and Group" transformation our models lag behind most frontier models. Also, for unseen compositions of three trained transformations our model also performs under par, indicating that our model cannot handle the increased complexity of applying three successive transformations. For transformations that are completely novel to our models (orange coloured bars), we see that our models do not perform as well as the frontier models. Frontier models may be trained on these transformations (i.e., reverse, replicate and shift) or may be able to generalize beyond their training to these simple transformation; however, it is clear that our models cannot solve completely novel transformations. 
\label{app:llmXtransformation}

\clearpage
\section{LLMs in few-shot setting versus MLC Copy 200 Model}
\label{app:llmfewshot}
\begin{figure}[ht]
    \centering
    \includegraphics[width=.8\textwidth]{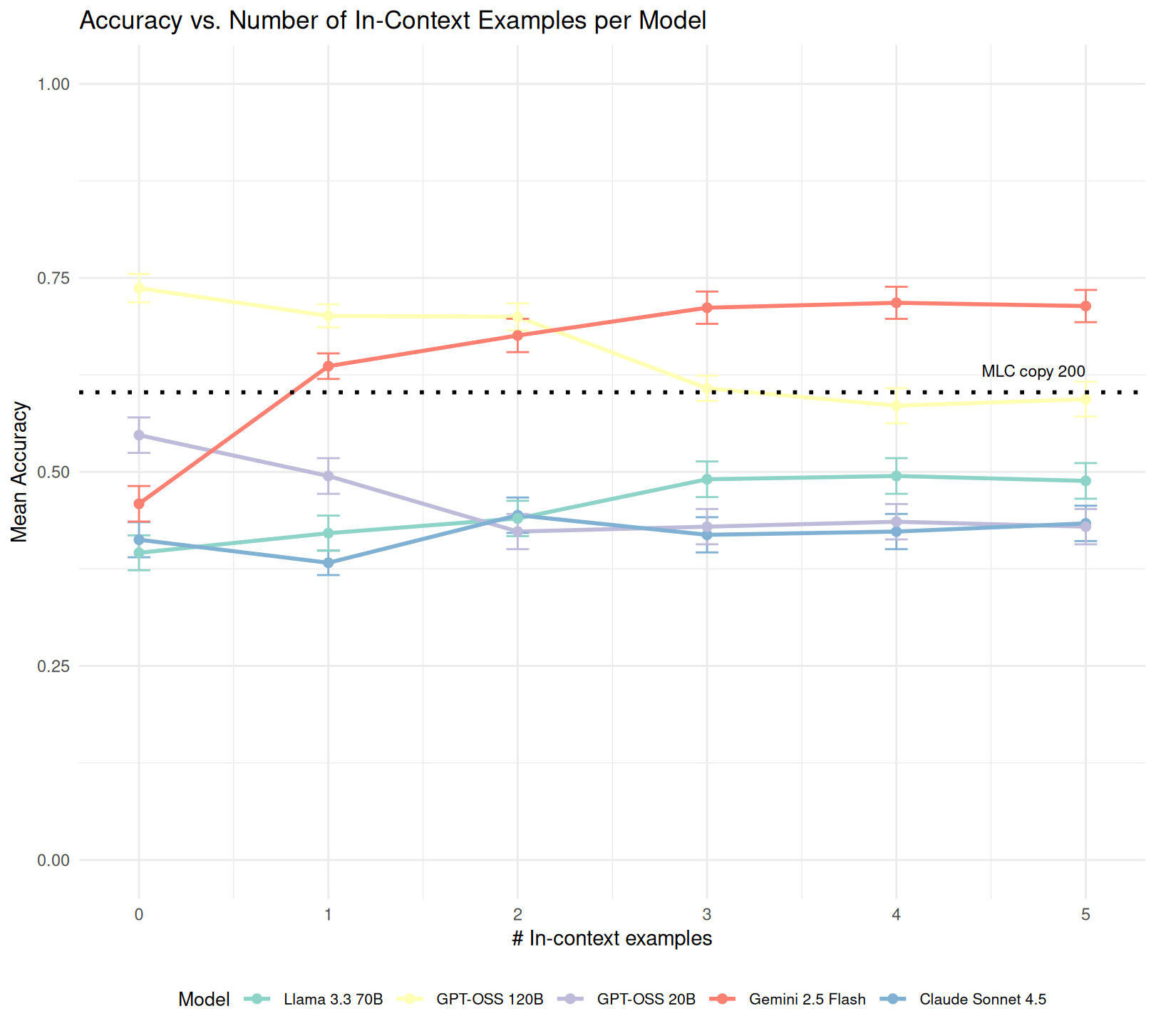}
    \caption{Frontier LLM  performance when given examples compared to our MLC copy 200 trained model.}
    \label{fig:llmfewshot}
\end{figure}

To mitigate the asymmetry in the comparison between our MLC models and LLMs, we examined whether LLMs benefited from a multi-shot set-up. As can be seen in Figure \ref{fig:llmfewshot}, the results are mixed when frontier LLMs (those from our original comparison that were still available) are given 1--5 in-context examples of how to solve each transformation using different permuted alphabets. Gemini Flash 2.5 and Llama 3.3 70B improved with examples, whereas GPT-OSS models' performance declined with two or more examples. %

\clearpage
\section{RASP-L Implementation of Predecessor Analogy}
\label{apx:raspl_impl}

\begin{lstlisting}[language=Python]
import np_rasp as rsp

# Input string
s = "abcdefghij|ghi>fhi|bcd"

# Convert chars to ascii ints
def tn(txt):
    return [ord(t) for t in txt]

## Layer 1
# Build selectors to identify the separators, the arrow, and to match elements of the input string to itself.
# Pipe selector: an attention weight matrix with 1 at the locations of pipe separators in each row and 0 everywhere else.
ppslct = rsp.select(k=tn(s), q=tn('|')*len(s), pred=rsp.equals, causal=False)
# Arrow selector: attention weight matrix with 1 at the location of the arrow in each row (>) nd 0 everywhere else.
gtslct = rsp.select(k=tn(s), q=tn('>')*len(s), pred=rsp.equals, causal=False)
# Same-as selector: attention weight matrix with, in each row, a 1 where the query matches the key and 0 everywhere else.
smslct = rsp.select(k=tn(s), q=tn(s),          pred=rsp.equals, causal=False) 

# Find the index, within the input string, of the first character after each separator, by aggregating over the relevant selectors.
# Finds the index of the character after the first separator
x1_ind = rsp.aggr(ppslct, v=rsp.indices(s)+1, reduction='min')
# Finds the index of the character after the arrow
x2_ind = rsp.aggr(gtslct, v=rsp.indices(s)+1, reduction='min')
# Finds the index of the character after the second separator
qy_ind = rsp.aggr(ppslct, v=rsp.indices(s)+1, reduction='max')

# Get first occurrence of each element of string
# Example output for s = "abcdefghij|ghi>fhi|bcd"
# [ 0  1  2  3  4  5  6  7  8  9 10  6  7  8 14  5  7  8 10  1  2  3]
fstind = rsp.aggr(smslct, v=rsp.indices(s),   reduction='min')

## Layer 2
# Locates the first character of the first part of the study example (x1), the first character of the second part of the study example (x2), and the first character of the query (qy) in the alphabet
# Locates x1. Attention weight matrix where each row of the matrix has a 1 at the index of x1 and 0 everywhere else.
x1slct = rsp.select(k=rsp.indices(fstind), q=x1_ind, pred=rsp.equals, causal=False)
# Locates x2. Attention weight matrix where each row of the matrix has a 1 at the index of x2 and 0 everywhere else.
x2slct = rsp.select(k=rsp.indices(fstind), q=x2_ind, pred=rsp.equals, causal=False)
# Locates qy. Attention weight matrix where each row of the matrix has a 1 at the index of qy and 0 everywhere else.
qyslct = rsp.select(k=rsp.indices(fstind), q=qy_ind, pred=rsp.equals, causal=False)

# Get the index of each of x1, x2, qy in the reference alphabet
x1_fst = rsp.aggr(x1slct, v=fstind, reduction='mean')
x2_fst = rsp.aggr(x2slct, v=fstind, reduction='mean')
qy_fst = rsp.aggr(qyslct, v=fstind, reduction='mean')

# Compute transformation of first element of query
# Corresponds to MLP operation
out_ind = x2_fst-x1_fst + qy_fst

## Layer 3
# Note that comparison is not exact because RASP-L does not yet support decoder and cross-attention
# Build component attention matrices id, o_slct, cpslct and combine using logical operations
# Identity matrix
id     = rsp.select(k=rsp.indices(s), q=rsp.indices(s), pred=rsp.equals, causal=True)
# Attention weight matrix where each row has a 1 for the transformed first element and 0 otherwise
o_slct = rsp.select(k=rsp.indices(s), q=out_ind,        pred=rsp.equals, causal=True)
# Attention weight matrix that selects all indices greater than the index of the first element of the query
cpslct = rsp.select(k=rsp.indices(s), q=qy_ind,         pred=rsp.gt,     causal=True)

# Build attention matrix which, for rows corresponding to the query (i.e., the last n rows), has a 1 at out_ind for the row corresponding to the first element of the query, and id for subsequent rows of the query, and 0 everywhere else. This is a different shape but comparable with last layer of decoder.
o_slct = (o_slct | cpslct ) & id

# We then aggregate with the original string (here, converted to ints since we have to take the 'mean' of this single element). Non-zero elements of out correspond to desired output.
out = rsp.aggr(o_slct, v=tn(s), default=0, reduction='mean')

# Viewing output NB not RASP, just convenience
# Outputs ['a', 'c', 'd']
out_view = [chr(c) for c in out if c != 0]
print(f"Nonzero output: {out_view}")
\end{lstlisting}

\clearpage
\subsection{Comparison between RASP-L Implementation and MLC Models}
\begin{figure}[h]
    \centering
    \includegraphics[width=\linewidth]{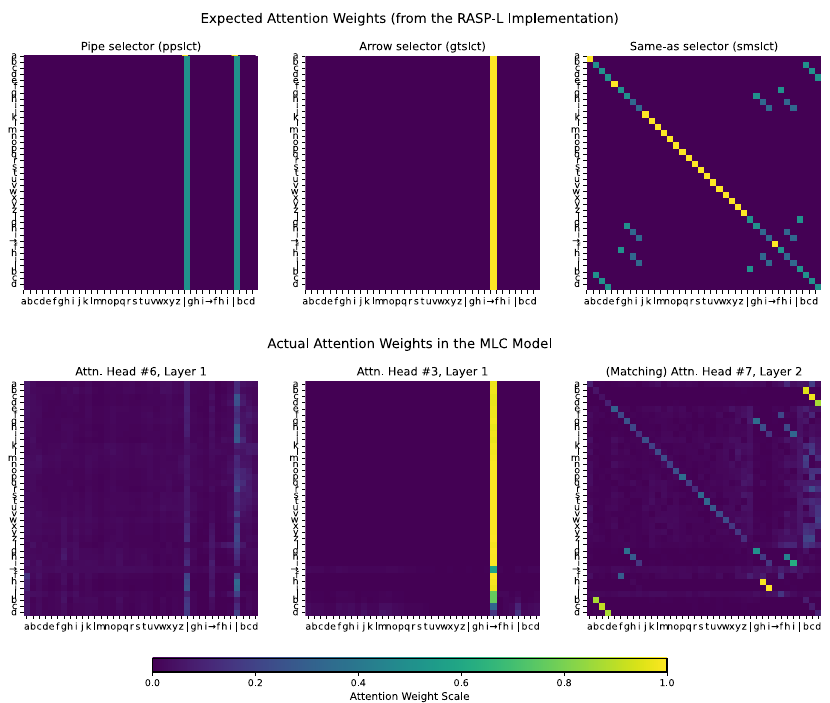}
    \caption{We observed clear parallels between our RASP-L implementation (top) and some of the attention heads in our MLC models (bottom). Most striking are the attention heads that correspond to the Arrow-selector (middle) and the Same-as selector (right). }
    \label{fig:rasp_attn_comparison}
\end{figure}

In addition to the experiments in the main text, we explicitly compared the attention heads of our models to the ones used in our RASP-L implementation for our running predecessor example. We find that for some attention heads, the attention patterns show strong similarities to the ones from our RASP implementation, most convincingly for the Arrow selector and the Same-as selector (Figure \ref{fig:rasp_attn_comparison}). In the RASP algorithm, the Arrow selector (top middle) is used to locate the arrow, so that elements of the study example can be located. The Same-as selector is used as part of the process for identifying where elements of the study example and query are located in the example alphabet.

\end{document}